\pdfoutput=1

\documentclass[11pt]{article}

\usepackage[final]{acl}

\usepackage{times}
\usepackage{latexsym}

\usepackage[T1]{fontenc}

\usepackage[utf8]{inputenc}

\usepackage{microtype}

\usepackage{array}
\usepackage{graphicx}
\usepackage{multirow}
\usepackage{xurl}
\usepackage{amsfonts}

\title{MuCoT: Multilingual Contrastive Training for \\ Question-Answering in Low-resource Languages}

\author{Gokul Karthik Kumar \quad Abhishek Singh Gehlot \\ \textbf{Sahal Shaji Mullappilly \quad Karthik Nandakumar}\\
 Mohamed Bin Zayed University of Artificial Intelligence (MBZUAI)\\
 Abu Dhabi, UAE\\
\tt\small{\{gokul.kumar, abhishek.gehlot, sahal.mullappilly , karthik.nandakumar\}@mbzuai.ac.ae} \\
}

\begin{document}
\maketitle
\begin{abstract}
Accuracy of English-language Question Answering (QA) systems has improved significantly in recent years with the advent of Transformer-based models (e.g., BERT). These models are pre-trained in a self-supervised fashion with a large English text corpus and further fine-tuned with a massive English QA dataset (e.g., SQuAD). However, QA datasets on such a scale are not available for most of the other languages. Multi-lingual BERT-based models (mBERT) are often used to transfer knowledge from high-resource languages to low-resource languages. Since these models are pre-trained with huge text corpora containing multiple languages, they typically learn language-agnostic embeddings for tokens from different languages. However, directly training an mBERT-based QA system for low-resource languages is challenging due to the paucity of training data. In this work, we augment the QA samples of the target language using translation and transliteration into other languages and use the augmented data to fine-tune an mBERT-based QA model, which is already pre-trained in English. Experiments on the Google ChAII dataset show that fine-tuning the mBERT model with translations from the same language family boosts the question-answering performance, whereas the performance degrades in the case of cross-language families. We further show that introducing a contrastive loss between the translated question-context feature pairs during the fine-tuning process, prevents such degradation with cross-lingual family translations and leads to marginal improvement. The code for this work is available at \url{https://github.com/gokulkarthik/mucot}.
\end{abstract}

\section{Introduction}

\begin{figure*}[!ht]
\centering
   \includegraphics[scale=0.525]{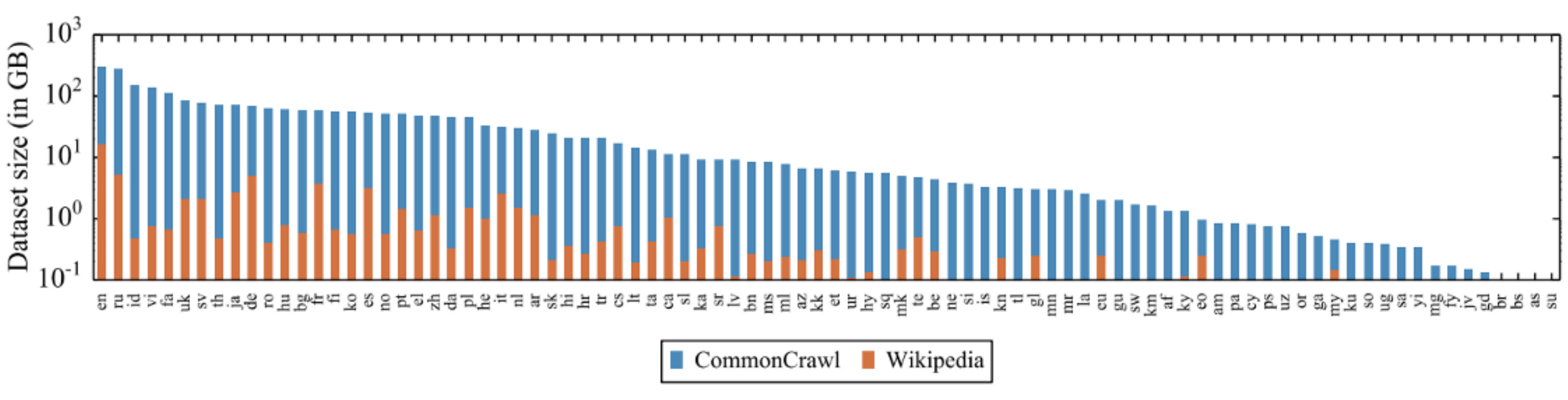}
   \caption{Amount of data in GB (log-scale) for the 88 languages that appear in both the Wiki-100 \cite{wiki100} corpus used for mBERT and XLM-100 \cite{conneau2020unsupervised}. None of the Indian languages feature among top-25 languages with the largest amount of data.}
   \label{fig:XLM_lang}
\end{figure*}

India has a population of 1.4 billion people speaking 447 languages and over 10,000 dialects, making it the country with the fourth-highest number of languages. However, Indian languages are highly under-represented on the Internet and Natural Language Processing (NLP) systems for Indian languages are in their nascency. Even state-of-the-art multilingual NLP systems perform sub-optimally on Indian languages \cite{ChAII}. This can be explained by the fact that multilingual language models are often jointly trained on 100+ languages and Indian languages constitute only a small fraction of their vocabulary and training data (as shown in Figure \ref{fig:XLM_lang}).

\begin{figure}[h]
\centering
   \includegraphics[scale=0.53]{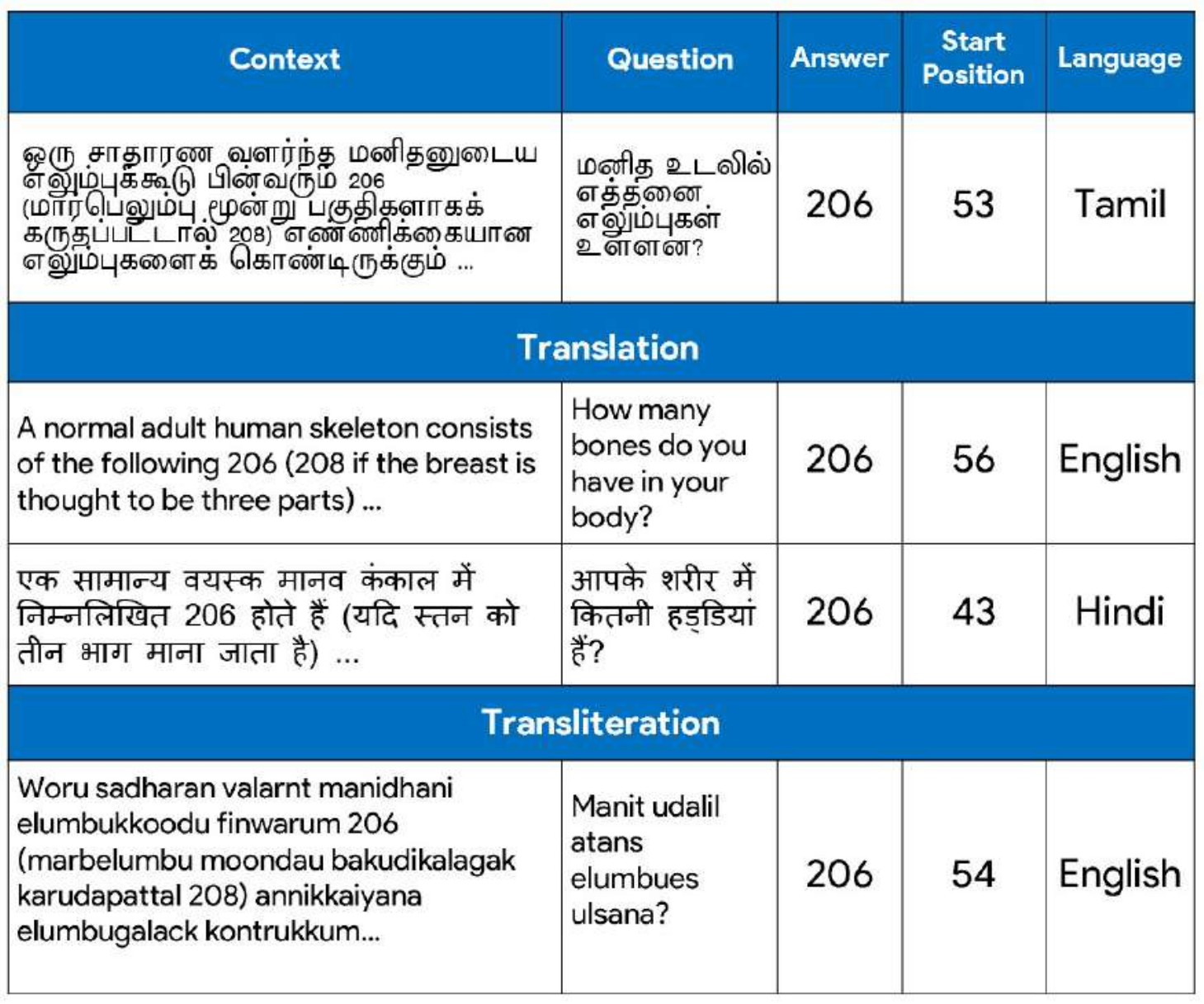}
   \caption{Example of a QA record from the ChAII QA dataset along with the translation and transliteration done on that record.}
   \label{fig:example}
\end{figure}

Machine learning models and tools have been proposed for many Natural Language Understanding tasks. In this work, we focus on Extractive Question-Answering (QA), where the goal is to localize the answer to a question within a large context (see Figure \ref{fig:example}). Specifically, we aim to develop a common multilingual question answering model for multiple Indian languages. A multilingual model has several advantages: (1) learning of cues across different languages, (2) a single model for many languages, and (3) avoiding dependency on English translation during inference. In this work, we start with a pre-trained multilingual Bidirectional Encoder Representations from Transformers (mBERT) model and further pre-train it with SQuAD \cite{rajpurkar2016SQuAD}, a large-scale question answering dataset in English. The resulting English-language mBERT-QA model is fine-tuned and evaluated for Indian languages Tamil and Hindi using the ChAII dataset \cite{ChAII}.

\label{sec:trans}
Fine-tuning the mBERT-QA model using only the training instances in the ChAII dataset is less effective because of the small number of training samples (1114 records with approximately two-thirds in Hindi and the rest in Tamil). To overcome this problem, we use translation and transliteration to other languages as a data augmentation strategy. The translation is the process of transforming the source content from one language to another, while the transliteration just involves modifying each word from the source content into another script. Both these operations are executed on the training dataset for the contexts, questions, and answers separately; then new locations of transformed answers in the transformed contexts are computed as shown in Figure \ref{fig:example}. Using translation and transliteration increases the size of the ChAII dataset manifold.

The choice of languages used for translation and transliteration is critical. \citet{kudugunta2019investigating} showed that languages under the same family have similar representations in multilingual models. Hence, we put together translations and transliterations from related languages within the same language family to achieve better performance. This will also help with better use of the vocabulary corpora from the low-resource languages. We also study the impact of translation and transliteration on languages outside the family of the target language. Since the cross-family language transfer degraded the QA performance, we introduce a contrastive loss {\citep{radford2021learning}} between the translated pairs to help retain or improve the original performance by encouraging the embeddings from all languages to be similar regardless of the family group. Thus, the contributions of the paper are three-fold:

\begin{itemize}
    \item We propose a three-stage training pipeline for question-answering in low-resource languages.
    \item We evaluate mBERT for question-answering in Tamil and Hindi with translations and transliterations as data augmentation techniques and show that same language family translations improve the performance. In contrast, we show that transliterations do not improve the QA performance on the ChAII dataset, regardless of the language family combinations.
    \item We propose a contrastive loss between the features of translated pairs to align the cross-family language representations.
\end{itemize}


\section{Related Work}
Bidirectional Encoder Representations from Transformers (BERT) \cite{devlin2018bert} is a deep learning model for general-purpose language representations. BERT is often used as the backbone model for several NLP tasks like semantic analysis, question answering, and named entity recognition. The bidirectional transformer used in BERT has a deeper sense of language context and generates intricate semantic feature representations. These representations are learned through a pre-training step using Next Sentence Prediction (NSP) and Masked Language Modelling (MLM) as pretext tasks and transferred to the downstream NLP tasks. The goal of the Next Sentence Prediction task is to identify whether the two input sentences are consecutive or not. In Masked Language Modelling, BERT is trained to predict randomly masked words in a sentence. The Transformer network receives a sequence of tokens as input and utilizes the attention mechanism to learn the contextual relationships between words in a text. These relationships can then be used to extract high-quality language features, which can be fine-tuned for applications like semantic analysis and question answering. Multi-lingual-BERT (mBERT) is a BERT model pre-trained using the Wikipedia text corpus \cite{wiki100} in more than 100 languages around the world. XLM-RoBERTa \cite{conneau2020unsupervised} scaled this idea with more than 2 terabytes of common crawl data.

Deep models such as Transformers rely heavily on the availability of a large amount of annotated data, which is available only for prominent languages like English, Russian, German or Spanish \cite{ponti2019modeling, joshi2020state}. For a majority of other languages with a minimal number of annotations, cross-lingual transfer learning \cite{prettenhofer2011cross, wan2011bi, ruder2019survey} has been proposed as a possible solution. This approach can transfer knowledge from the annotation-rich source language to low-resource or zero-resource target languages. Furthermore, multilingual models \cite{lewis2019mlqa, clark2020tydi} can be used to mitigate the data scarcity problem. For example, LASER \cite{artetxe2019massively} used a bidirectional LSTM  \cite{10.1162/neco.1997.9.8.1735} encoder with a byte pair encoding vocabulary shared between languages. This work showed that joint training of multiple languages helped to improve the model performance for low-resource languages. LaBSE \cite{feng2020language} used the mBERT \cite{devlin2018bert} encoder pre-trained with masked language modelling and translation language modelling \cite{lample2019cross} tasks. It attempted to optimize the dual encoder translation ranking \cite{guo-etal-2018-effective} loss during pre-training to achieve similar embedding for the same text in different languages.

The work of \citet{bornea2020multilingual} showed that large pre-trained multilingual models are not enough for question-answering in under-represented languages and presented several novel strategies to improve the performance of mBERT with translations. This work achieved language-independent embeddings, which improved the cross-lingual transfer performance with additional pre-training on adversarial tasks. It also introduced a Language Arbitration Framework (LAF), which consolidated the embedding representations across languages using properties of translation. Cross-lingual manifold mixup (X-Mixup) \cite{yang2021enhancing} achieved better cross-lingual transfer by calibrating the representation discrepancy, which resulted in a compromised representation for target languages. It was shown that the multilingual pre-training process can be improved by implementing X-Mixup on parallel data. Contrastive Language-Image pre-training (CLIP) \cite{radford2021learning} introduced an efficient way to learn scalable image representations with natural language supervision. Drawing inspiration from ConVIRT \cite{zhang2020contrastive}, CLIP used a contrastive objective that maximizes the cosine similarity of the correct pairs of images and text, while minimizing the same for incorrect pairs.

Building upon the work of \cite{bornea2020multilingual}, we show that translations of a small-scale dataset into cross-family languages could degrade the QA performance. To overcome this problem, we propose multilingual contrastive training to encourage cross-lingual invariance. Our approach is relatively simpler compared to adversarial training and LAF used in \citet{bornea2020multilingual}. Though the proposed contrastive loss has a similar objective to the pre-training loss in \cite{guo-etal-2018-effective}, there are subtle differences because we use it in multi-task learning setup along with the original task loss for fine-tuning.

\section{Methodology}

\begin{figure*}[ht]
\centering
   \includegraphics[width=0.9\linewidth]{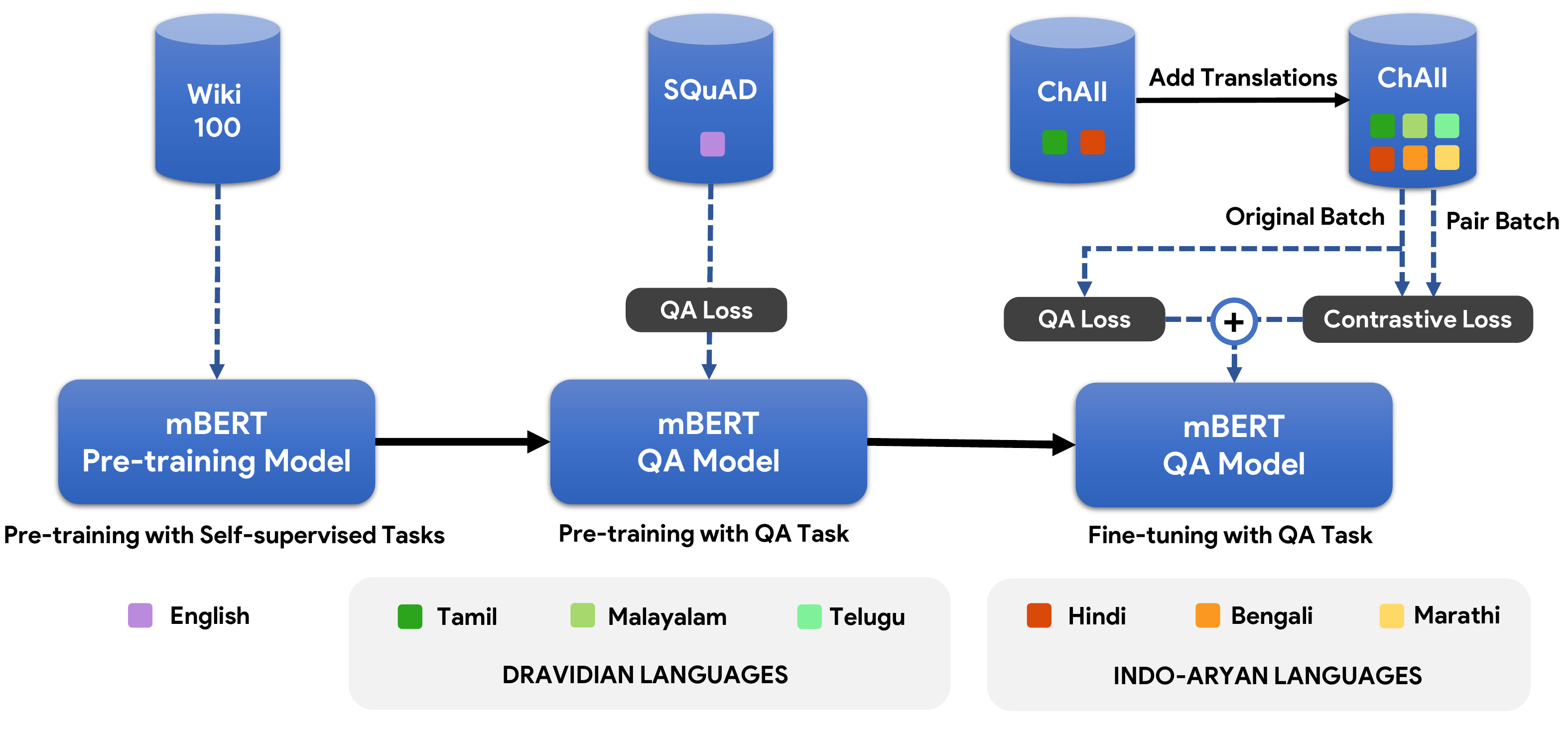}
   \caption{Proposed training pipeline of MuCoT for question answering in low resource languages Tamil and Hindi.}
   \label{fig:framework}
\end{figure*}

\subsection{Data Representation and Baseline Model}    

We adopt the standard data representations that are commonly used in Transformer-based question-answering models. We use the same word-piece Tokenizer of mBERT to tokenize the concatenated input of question-context pairs. For the question answering task, the context is usually very long. In some NLP applications, truncating the input text is a viable choice because it leads to only loss of information. But in the extractive question answering task, removing part of the context may result in loss of answer as well. To overcome this challenge, we follow the popular approach of splitting the long context into parts that fit into the model and regulate this splitting using an additional hyper-parameter called 'max length'. Moreover, to cover for cases where the answer might be distributed over multiple splits of the context, an overlap factor is introduced, which in turn is controlled by another hyper-parameter 'doc stride'.

Our baseline is the mBERT model \cite{devlin2018bert}, which is pre-trained using pretext tasks like Masked Language Modelling and Next Sentence Prediction on a multilingual text corpus that includes our target languages, Hindi and Tamil. The default output head of mBERT is replaced with the head for the question-answering task. This is done by adding separate output heads for classifying the start and end positions as shown in \citet{devlin2018bert}.

\subsection{Proposed Framework for Effective Cross-lingual Transfer}

We propose a three-stage pipeline called Multilingual Constrative Training (MuCoT) to effectively train the mBERT model for question-answering in low-resource languages. An illustration of this pipeline for two low-resource languages, namely Tamil and Hindi, is shown in Figure \ref{fig:framework}. The first stage is pre-training the baseline multilingual model (mBERT). The second stage involves pre-training the QA head using the large-scale dataset(s) in high resource language(s). In Figure \ref{fig:framework}, English is considered the high-resource language and SQuAD \cite{rajpurkar2016SQuAD} dataset is used to pre-train the QA head and obtain the mBERT-QA model. The final stage involves fine-tuning the mBERT-QA model using both original and augmented samples from the target low-resource languages. In this work, ChAII \cite{ChAII} dataset is used for obtaining training samples in Tamil and Hindi.

Since SQuAD \cite{rajpurkar2016SQuAD} and ChAII \cite{ChAII} datasets have similar Wikipedia\footnote{\url{https://www.wikipedia.org/}} style contexts, it is possible to train a multilingual QA model jointly using both datasets. However, to take advantage of the engineering and training efforts of publicly available models, we sequentially use both these datasets. After obtaining the mBERT-QA model pre-trained for the English language QA task, we fine-tune it on the ChAII dataset using the following loss function.

\begin{equation}
   L_{total} = L_{task} + w_{contrastive} * L_{contrastive},
\end{equation}

\noindent where $L_{task}$ and $L_{contrastive}$ are the QA task loss and multilingual contrastive loss, respectively, $L_{total}$ is the total loss, and $w_{contrastive}$ is the relative weight assigned to the contrastive loss. Note that fine-tuning using only the QA task loss is often not sufficient to achieve good performance, especially if the dataset used for fine-tuning is small. To mitigate this problem, we translate the training samples into other languages and use both original and translated samples for fine-tuning. While this approach works well for translations into other languages within the same language family, it leads to sub-optimal performance in the case of cross-family language translations, due to divergence in the representations across language families. To solve this issue, we introduce the multi-lingual contrastive loss $L_{contrastive}$.
 
\begin{figure*}[ht]
\centering
   \includegraphics[width=0.9\linewidth]{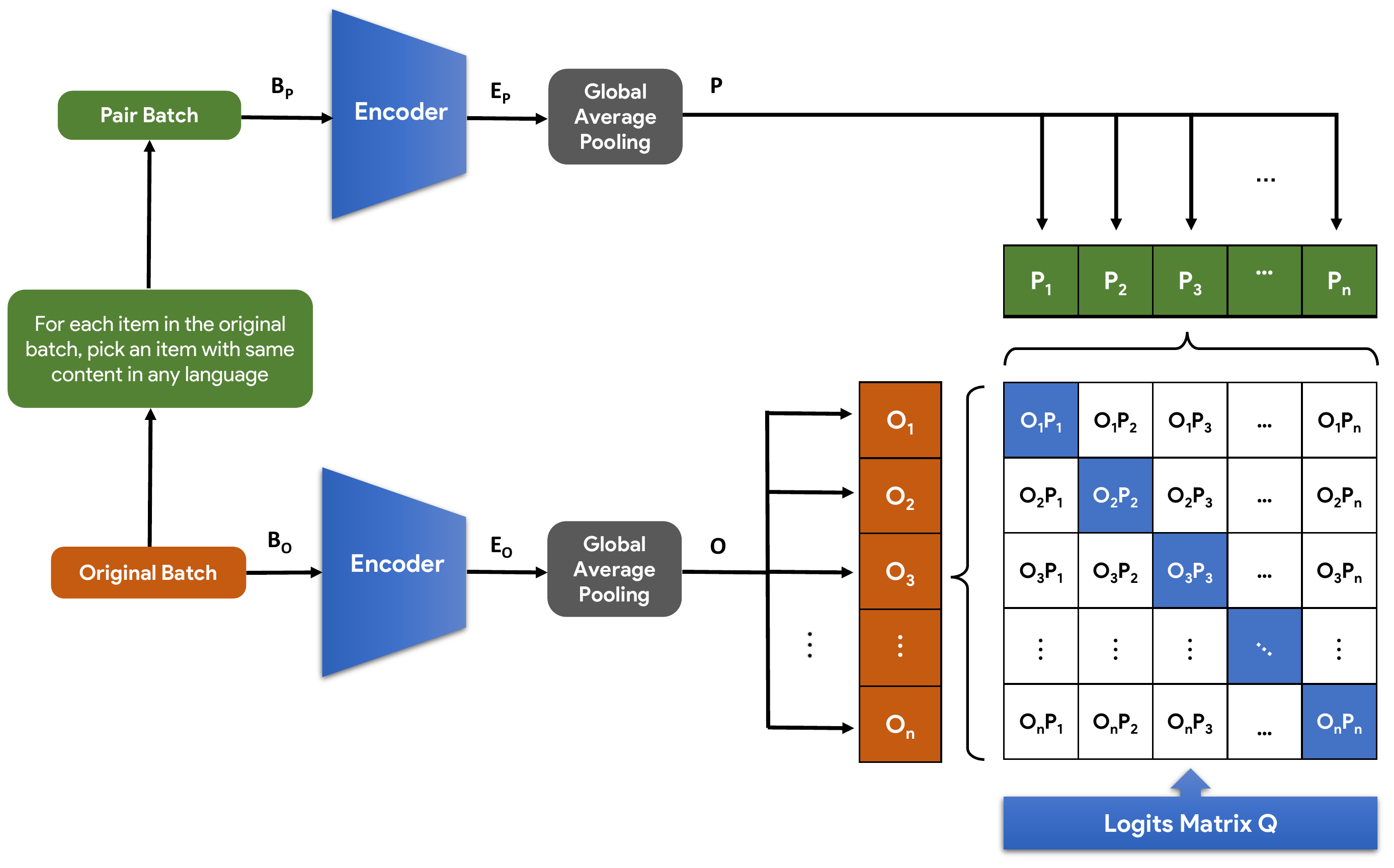}
   \caption{Logits matrix computation for the input to contrastive loss, similar to CLIP \cite{radford2021learning}}
   \label{fig:contrastive}
\end{figure*}

\subsection{Multilingual Contrastive Loss}
\label{sec:contrastive}

During fine-tuning, for each data point in the original batch ($B_{o}$) of size $n$, we pick one of its corresponding translations uniformly at random and form a translated batch ($B_{p}$) of the same size $n$. It is important to note that  $B_{o}$ itself is taken from the combined dataset of source instances and translated instances. The two batches that form a pair are denoted as original batch and pair batch, respectively, in Figure \ref{fig:contrastive}. We use the same mBERT network up to a specific layer as our encoder ($enc$) to transform $B_{o}$ and $B_{p}$ to get the embeddings, $E_{o}, E_{p} \in \mathbb{R}^{n*t*d} $, respectively. Then, we apply a global average pooling ($gap$) operation to aggregate the vector representations of $t$ tokens into a single vector representation of dimension $d$ for each instance in each batch. This will result in the aggregated embeddings $O, P \in \mathbb{R}^{n*d}$ for $B_{o}$ and $B_{p}$, respectively. With these $n$ feature vectors in the original and the translated batch, we follow the CLIP \cite{radford2021learning} approach and compute the contrastive loss using the cross-entropy loss ($L_{ce}$). Specifically, we multiply the matrices $O$ and $P^T$ to get the logits matrix $Q \in \mathbb{R}^{n*n}$. Then, we apply the cross-entropy loss $L_{ce}$ row-wise and column-wise to the logits matrix $Q$, with its diagonal locations as original classes for each row and column, respectively.

\begin{equation}
    O = gap(enc(B_{o})),
\end{equation}
\begin{equation}
    P = gap(enc(B_{p})),
\end{equation}
\begin{equation}
    Q = OP^T,
\end{equation}
\begin{equation}
    L_{contrastive} = \frac{L_{ce}^{row} (Q) + L_{ce}^{column} (Q)}{2}.
\end{equation}

\section{Experimental Results}

\subsection{Datasets}
In our experiments, we use ChAII \cite{ChAII} question-answering dataset for fine-tuning and evaluation. This dataset was recently released by Google Research India and has 1,114 records of context, question, answer, and its corresponding start position in the context for Tamil and Hindi languages. Hindi is represented predominantly in the dataset with nearly two-thirds of the records. As the ChAII dataset has been published as part of an ongoing Kaggle competition \cite{ChAII}, the complete test dataset has not been disclosed to the public. Hence, we have used Scikit-learn's train\_test\_split method with a test size of $100$, stratified on language and with a random seed of $0$, to get the $test$ split from the training data. Similarly, we applied the same method over the filtered $train$ split to get the $validation$ split of 100 samples. We also use the translations and transliterations of this training split as augmented samples for fine-tuning the QA model.

Stanford Question Answering Dataset (SQuAD) \cite{rajpurkar2016SQuAD} is the most popular question-answering dataset in English. This dataset had been crowdsourced to form 100K records of answerable question-answer pairs along with the context. This dataset is used to pre-train the QA head added to the pre-trained mBERT model, which is subsequently fine-tuned using the ChAII dataset.

\subsection{Translation and Transliteration Details}
We use AI4Bharat's IndicTrans\footnote{\url{https://indicnlp.ai4bharat.org/indic-trans/}} \cite{ramesh2021samanantar} for translation, which is a Transformer-4X model trained on $Samanantar$ dataset \cite{ramesh2021samanantar}. In IndicTrans, translation can be done from Indian languages to English and vice versa. Available Indian languages include Assamese, Bengali, Gujarati, Hindi, Kannada, Malayalam, Marathi, Oriya, Punjabi, Tamil, and Telugu. At first, we translate the ChAII dataset from Hindi and Tamil to English and then to Bengali, Marathi, Malayalam, and Telugu. In the FLORES devset benchmark \cite{goyal2021flores}, the BLEU scores of IndicTrans for translating Hindi and Tamil to English are 37.9 and 28.6, respectively. The scores for translating English to Bengali, Marathi, Malayalam, and Telugu are 20.3, 16.1, 16.3, and 22.0, respectively. We were not able to translate nearly 500 of the ChAII instances to English as the automatic search for the translated answers in the translated contexts failed. This happened because the same word got translated differently in the context and the answer. For the same reason, we lost nearly another 200 instances when translating from English to other Indian languages.

For transliteration, we use the open-source Indic-trans transliteration module\footnote{\url{https://indic-trans.readthedocs.io/en/latest/index.html}} \cite{Bhat:2014:ISS:2824864.2824872}, which is available for many Indian language scripts including English and Urdu. Here, we directly transliterate from Hindi and Tamil to Bengali, Marathi, Malayalam, and Telugu.

\subsection{Model Training Details}
We used mBERT\footnote{\url{https://huggingface.co/bert-base-multilingual-cased}} as our baseline model. It is modified for the question-answering task by replacing the output head using HuggingFace's auto model. At first, we evaluated this model after directly fine-tuning on the train split of the ChAII dataset. Then, we introduced intermediate SQuAD pre-training and fine-tuned on the train split of the ChAII dataset with and without translations or transliterations.  The hyperparameter settings listed in Table \ref{tab:hyperparam} are used for all the experiments. We have experimented with different levels of mBERT layers to compute the contrastive loss. Layer 3 performed consistently well compared to the initial layer 1 and the deeper layers such as 5. Initially, we used contrastive training for all the steps. However, forcing the model to learn exact representations across languages could make the model forget the task-specific patterns learned with intermediate pre-training on a large-scale dataset. Hence, we applied the contrastive loss only for training steps that are a multiple of 500 and picked the best one. Other hyperparameters are tuned based on a standard search over multiple choices.

\begin{table}[h]
\begin{center}
\begin{tabular}{c c}
\hline
\textbf{Hyperparameter}         & \textbf{Value}\\ 
\hline
Maximum feature length & 128                                                 \\
Document stride        &  384                                                   \\
Batch size             & 16                                                   \\
Maximum optimization steps         & 5000                                                  \\
Learning rate          & 0.00003                                           \\
Weight decay           & 0.01              \\
Contrastive loss layer & 3 \\
Contrastive loss weight & 0.05 \\
Maximum contrastive steps & 1000 \\
\hline
\end{tabular}
\caption{Hyperparameter configuration of all the models for fine-tuning on ChAII dataset}
\label{tab:hyperparam}
\end{center}
\end{table}

\begin{table*}[!hbt]
\centering
\begin{tabular}{|c|c|c|cc|cc|cc|}
\hline
\textbf{SQuAD pre-training}            & \textbf{No}          & \textbf{Yes}         & \multicolumn{2}{c|}{\textbf{Yes}}                & \multicolumn{2}{c|}{\textbf{Yes}}                 & \multicolumn{2}{c|}{\textbf{Yes}}             \\
\hline
\textbf{Translations}                  & \textbf{No}          & \textbf{No}          & \multicolumn{2}{c|}{\textbf{Dravidian (ml, te)}} & \multicolumn{2}{c|}{\textbf{Indo-Aryan (bn, mr)}} & \multicolumn{2}{c|}{\textbf{All languages}}   \\
\hline
\textit{\textbf{Contrastive Training}} & \textit{\textbf{No}} & \textit{\textbf{No}} & \textit{\textbf{No}}   & \textit{\textbf{Yes}}  & \textit{\textbf{No}}   & \textit{\textbf{Yes}}   & \textit{\textbf{No}} & \textit{\textbf{Yes}} \\
\hline
\textbf{Overall}                       & 0.44                 & 0.5                  & 0.49                   & \textbf{0.53}          & 0.51                   & 0.52                    & 0.49                 & 0.52                  \\
\textbf{Hindi}                         & 0.47                 & 0.57                 & 0.52                   & 0.57                   & \textbf{0.59}          & 0.58                    & 0.54                 & 0.57                  \\
\textbf{Tamil}                         & 0.39                 & 0.37                 & 0.44                   & \textbf{0.45}          & 0.35                   & 0.4                     & 0.39                 & 0.41 \\
\hline
\end{tabular}
\caption{Jaccard scores with translation used as augmentation in different training settings. ml, te, bn, and mr denote Malayalam, Telugu, Bengali, and Marathi, respectively.}
\label{table:trans}
\end{table*}

\begin{table*}[!hbt]
\centering
\begin{tabular}{|c|c|c|cc|cc|cc|}
\hline
\textbf{SQuAD pre-training}            & \textbf{No}          & \textbf{Yes}         & \multicolumn{2}{c|}{\textbf{Yes}}                & \multicolumn{2}{c|}{\textbf{Yes}}                 & \multicolumn{2}{c|}{\textbf{Yes}}             \\
\hline
\textbf{Transliterations}              & \textbf{No}          & \textbf{No}          & \multicolumn{2}{c|}{\textbf{Dravidian (ml, te)}} & \multicolumn{2}{c|}{\textbf{Indo-Aryan (bn, mr)}} & \multicolumn{2}{c|}{\textbf{All languages}}   \\
\hline
\textit{\textbf{Contrastive Training}} & \textit{\textbf{No}} & \textit{\textbf{No}} & \textit{\textbf{No}}   & \textit{\textbf{Yes}}  & \textit{\textbf{No}}   & \textit{\textbf{Yes}}   & \textit{\textbf{No}} & \textit{\textbf{Yes}} \\
\hline
\textbf{Overall}                       & 0.44                 & 0.5                  & 0.5                    & 0.49                   & \textbf{0.53}          & 0.47                    & 0.49                 & 0.46                  \\
\textbf{Hindi}                         & 0.47                 & 0.57                 & 0.52                   & 0.55                   & \textbf{0.56}          & 0.53                    & 0.52                 & 0.53                  \\
\textbf{Tamil}                         & 0.39                 & 0.37                 & \textbf{0.45}          & 0.36                   & 0.44                   & 0.36                    & 0.44                 & 0.32       \\
\hline
\end{tabular}
\caption{Jaccard scores with transliteration used as augmentation in different training settings. ml, te, bn, and mr denote Malayalam, Telugu, Bengali, and Marathi, respectively.}
\label{table:tlit}
\end{table*}

\subsection{Evaluation Metric}

Given the noisy nature of the ChAII dataset, we employed the Jaccard score as the evaluation metric. Jaccard similarity coefficient is widely used for determining similarity between sets/intervals and is defined as $J (A, B) = \frac {|A \cap B|} {|A \cup B|}$. Here, $A$ and $B$ are sets/intervals, and $\cap$ and $\cup$ represent intersection and union, respectively. We compute the evaluation metric for the overall test split as well as for individual language test sets in intervals of 500 optimization steps. For each experiment, we pick the model at a specific optimization step that gives the best overall Jaccard score and reports its performance.

\subsection{Performance}

As shown in Tables \ref{table:trans} and \ref{table:tlit}, translation and transliteration affect the performance in different ways. While some data is lost during the translation process due to failed automatic search of translated text in the translated context. transliteration does not cause any such loss. However, to ensure a fair comparison, records lost during translation are dropped from transliterated testing as well. Note that we use the same hyper-parameters from Table \ref{tab:hyperparam} for evaluating the models and later stages with additional augmentation and contrastive training.

First, we observe from Table \ref{table:trans} that just having intermediate SQuAD pre-training in English, improves the overall Jaccard score significantly from $0.44$ to $0.5$. Furthermore, we fine-tune by dividing translated and transliterated data into Indo-Aryan and Dravidian language families to study how translated and transliterated pairs serve as supervised cross-lingual signals when languages share semantics and structure \cite{mikolov2013exploiting}. Although transliteration improves the Jaccard scores in certain cases compared to the baseline, the trend is not consistent. Moreover, contrastive training does not help in the case of transliteration as shown in Table \ref{table:tlit}. This could be because the QA model is pre-trained only with regular text and not with transliteration style text.

From Table \ref{table:trans}, we observe that grouped translated data in the same language family helps in improving performance. The translated Indo-Aryan data (Bengali and Marathi) increases the Jaccard score of Hindi answers to $0.59$ from $0.57$. Similarly, Dravidian language data (Telugu and Malayalam) significantly increase the Jaccard similarity of Tamil answers from $0.37$ to $0.44$. At the same time, the overall Jaccard score did not change much because of the degradation in cross-family language performance. Interestingly, we could observe in Table \ref{table:trans} that the contrastive training helps in preventing such degradation and improves the overall score by encouraging similar representations between languages from across families.

\section{Conclusion and Future Work}
With Internet usage expanding every day, there is an increasing need to develop better NLP models for a variety of downstream tasks in vernacular languages. As most of these languages do not have labeled resources that are sufficient to train stand-alone modern deep learning models, we need to rely on multilingual models and enhance them. Our work is a step in this direction and is an attempt to understand and evaluate the impact of cross-lingual knowledge transfer through pre-training and fine-tuning. We utilize modern open-source deep learning models to translate the ChAII dataset into different languages from two language families namely, Dravidian, and Indo-Aryan, and use them to improve the question-answering performance. Our analysis shows an effective way to pick languages for translation, which can be used for fine-tuning. We also showed that introducing a contrastive loss with the original task training loss increases the performance even for cross-family languages.

Despite the inclusion of translations and contrastive loss, we observed that there is only a marginal improvement in the QA performance. This can be attributed to the smaller size of the ChAII dataset with 1114 instances (Tamil and Hindi combined; Train, Validation, and Test combined), which is clearly insufficient to fine-tune a 177M parameter model. Hence, the proposed techniques have to be evaluated on other larger datasets as well as using other multilingual models like XLM-RoBERTa \cite{conneau2020unsupervised}, Distill-mBERT \cite{sanh2019distilbert}, MURIL \cite{khanuja2021muril} and Indic-BERT \cite{kakwani2020indicnlpsuite}. We hope that the proposed techniques will motivate further research in this field, including exploration of the same phenomenon of cross-lingual transfer in other language families and multilingual tasks.


\bibliographystyle{acl_natbib}
\bibliography{main}

\end{document}